%% file: main.tex
\documentclass[letterpaper]{article} 
\usepackage{aaai2026}  
\usepackage{times}  
\usepackage{helvet}  
\usepackage{courier}  
\usepackage[hyphens]{url}  
\usepackage{graphicx} 
\usepackage{natbib}  
\usepackage{caption} 
\usepackage{algorithm}

\usepackage{algpseudocode}
\usepackage{amsmath}
\usepackage{amsfonts}
\usepackage{amssymb}
\usepackage{booktabs}
\usepackage{multirow}
\usepackage{newfloat}
\usepackage{listings}
\DeclareCaptionStyle{ruled}{labelfont=normalfont,labelsep=colon,strut=off} 
\floatstyle{ruled}
\newfloat{listing}{tb}{lst}{}
\floatname{listing}{Listing}
\title{Distillation-Guided Structural Transfer for Continual Learning Beyond Sparse Distributed Memory}
\author{
    Huiyan Xue\textsuperscript{\rm 1}\thanks{These authors contributed equally.},\,
    Xuming Ran\textsuperscript{\rm 2}\footnotemark[1],\,
    Yaxin Li\textsuperscript{\rm 1},\,
    Qi Xu\textsuperscript{\rm 1}\thanks{Corresponding authors.},\,
    Enhui Li\textsuperscript{\rm 1},\,
    Yi Xu\textsuperscript{\rm 1},\,
    Qiang Zhang\textsuperscript{\rm 1}\footnotemark[2]
}
\affiliations{
    \textsuperscript{\rm 1}School of Computer Science and Technology, Dalian University of Technology, China\\
    \textsuperscript{\rm 2}National University of Singapore, Singapore\\
    xuqi@dlut.edu.cn, zhangq@dlut.edu.cn
}

\usepackage{bibentry}

\begin{document}

\maketitle

\begin{abstract}
Sparse neural systems are gaining traction for efficient continual learning due to their modularity and low interference. Architectures like Sparse Distributed Memory Multi-Layer Perceptrons (SDMLP) construct task-specific subnetworks via Top-K activation and have shown resilience against catastrophic forgetting. However, their rigid modularity poses two fundamental challenges: (1) the isolation of sparse subnetworks severely limits cross-task knowledge reuse; and (2) increased sparsity reduces interference but often degrades performance due to constrained feature sharing.
We propose \textbf{S}elective \textbf{S}ubnetwork \textbf{D}istillation (\textbf{SSD}), a structurally guided continual learning framework that treats distillation not as a regularizer, but as a topology-aligned information conduit. By identifying neurons with high activation frequency, SSD selectively distills knowledge within previous Top-K subnetworks and output logits—without requiring replay or task labels—preserving both sparsity and functional specialization.
Unlike conventional distillation, SSD operates under hard modular constraints and enables structural realignment without altering the sparse architecture.
While our method is validated on SDMLP, its structure-aligned mechanism has the potential to generalize to other sparse networks as a plug-in module for promoting representation sharing.
Comprehensive experiments on Split CIFAR-10, CIFAR-100, and MNIST demonstrate that SSD improves accuracy, retention, and manifold coverage, offering a structurally grounded solution to sparse continual learning.
\end{abstract}


\section{Introduction}
In Continual Learning (CL), models are required to sequentially acquire new tasks while avoiding the forgetting of previously learned knowledge~\cite{french1999catastrophic}. However, artificial neural networks often suffer from catastrophic forgetting during sequential learning, where representations of earlier tasks are overwritten by new ones, severely degrading long-term performance ~\cite{ DeLange2021Continual, aljundi2019task}.

Inspired by the sparse activation mechanisms in biological neural systems ~\cite{kanerva1988sparse}, Sparse Distributed Memory (SDM)-based improvements to Multi-Layer Perceptrons (MLPs) have been proposed ~\cite{li2023sparse, sung2021fixedmask}. These methods leverage the Top-$K$ activation function to retain only the most active neurons in each layer, dynamically forming input-dependent subnetworks ~\cite{frankle2019lottery, evci2022lottery}. This sparsity reduces task interference at the parameter level, enabling continual learning without replay ~\cite{mai2022mir}. Nevertheless, the subnetworks activated for different tasks in SDMLP are typically disjoint and structurally isolated, which restricts cross-task knowledge transfer and limits the reuse of prior representations. This independence degrades generalization in dynamic environments, particularly under high sparsity, where the model struggles with representation bottlenecks and a trade-off emerges between accuracy and sparsity.

To address this structural bottleneck, we propose \textbf{S}elective \textbf{S}ubnetwork \textbf{D}istillation (\textbf{SSD}), which enhances continual learning through structure-aware knowledge transfer. Instead of relying on global distillation or task labels, SSD selects Top-$n$ neurons based on their historical activation frequency to perform layer-wise distillation and construct hierarchical connection paths between corresponding subnetworks. This structural routing mechanism, inspired by selective activation in neurophysiology \cite{olshausen2023sparse}, effectively establishes inter-subnetwork knowledge channels, enabling structured sharing and transfer \cite{DBLP:journals/corr/abs-2410-05899}. SSD surpasses SDMLP in both structural design and functional capacity. It addresses SDMLP’s core limitations: disconnected subnetworks and insufficient knowledge reuse, and the trade-off between accuracy and sparsity, thereby providing a structural enhancement~\cite{wang2022sparcl} strategy tailored for sparse continual learning.

In contrast to conventional knowledge distillation techniques that operate in the output space via loss functions \cite{hinton2015distilling, gou2021kdsurvey}, SSD functions not as a loss adjustment but as an architectural mechanism. It explicitly connects sparse task-specific subnetworks through inter-layer paths, enabling transferable representation sharing at the architectural level. This structural advantage allows SSD to function without sample replay or task labels, making it well-suited for more complex and dynamic continual learning scenarios \cite{bhat2022sscl, hafez2024taskagnosticrl}.

The primary contributions of this work are summarized as follows:
\begin{itemize}
    \item A systematic analysis of structural bottlenecks in sparse continual learning frameworks such as SDMLP, particularly in terms of knowledge transfer and representational capacity ~\cite{DeLange2021Continual, li2024kdsurvey};
    \item The proposal of \textbf{SSD}, a structure-enhanced method that performs layer-wise selective distillation based on neuron activation and subnetwork connectivity, achieving cross-task knowledge transfer without relying on replay or task labels;
    \item Extensive experiments on standard continual learning benchmarks including Split CIFAR-10, CIFAR-100, and MNIST, where SSD achieves significant improvements in classification accuracy and knowledge retention, validating its effectiveness and generality within sparse continual learning settings.
\end{itemize}
 
The remainder of this paper is organized as follows: Section 2 reviews related work on sparse continual learning and knowledge distillation; Section 3 introduces the background and mechanisms of Sparse Distributed Memory and SDMLP; Section 4 presents the design and training strategy of the proposed Selective Subnetwork Distillation; Section 5 reports experimental results and ablation studies; and Section 6 discusses the method’s advantages, limitations, and potential extensions, followed by the conclusion.

\section{Related Work}
\subsection{Sparse Subnetwork-Based Continual Learning Methods}
  
Continual learning techniques can generally be categorized into three major paradigms: architectural approaches \cite{goodfellow2013empirical,goodfellow2014str}, regularization-based approaches \cite{smith2022always,kirkpatrick2017overcoming,zenke2017synaptic,aljundi2018memory}, and replay-based methods \cite{DeLange2021Continual,hsu2018re}. Among these, architectural approaches, which design task-specific network structures, often leverage sparse subnetwork strategies to minimize inter-task interference.\cite{aljundi2019task,xu2018reinforced,mallya2018piggyback,schwarz2021progress,smith2022always}.  

Specifically, some methods use weight masking or dynamic routing to partition pre-trained networks into task-specific subnetworks. For instance, Piggyback~\cite{mallya2018piggyback} learns binary masks to select substructures on fixed weights for new tasks, Supermasks in Superposition~\cite{wortsman2020supermasks} identifies sparse subnetworks without altering weights, and PackNet~\cite{mallya2018packnet} prunes redundant weights to allocate capacity for subsequent tasks. These approaches typically require explicit task boundaries and additional memory for masks or weight snapshots. Alternatively, sparse activation mechanisms limit active neurons per layer to dynamically form task-specific subnetworks, well-suited for task-free continual learning. However, these methods often lack effective knowledge transfer across tasks, limiting long-term generalization.

\subsection{Knowledge Distillation in Continual Learning}
Knowledge distillation (KD)~\cite{hinton2015distilling} has been widely applied to continual learning. 
Following the taxonomy in~\cite{li2024kdsurvey}, KD-based continual learning approaches can be categorized into three paradigms: regularization-based, data replay-based, and feature replay-based distillation. 

Regularization-based methods constrain the current model’s predictions to match those of the previous model, enabling knowledge retention without data replay, such as Learning without Forgetting (LwF)~\cite{li2018lwf} using logits distillation and Learning without Memorizing (LwM)~\cite{dhar2019learning} employing attention-based feature alignment. Data replay-based methods store or generate past task samples, like iCaRL~\cite{rebuffi2017icarl} combining exemplars with distillation loss, Deep Generative Replay (DGR)~\cite{shin2017dgr} synthesizing data, and SparCL~\cite{wang2022sparcl} optimize replay efficiency for resource-limited devices. Feature replay-based methods reconstruct old task feature spaces to avoid raw data storage, as in GFR~\cite{belouadah2020gfr} and PLOP~\cite{douillard2021plop} extends this idea to dense prediction tasks such as semantic segmentation.

These methods significantly enhance continual learning but focus on dense architectures and often rely on replay, limiting sparse network applications. Our Sparse Subnetwork Distillation (SSD) balances sparsity and cross-task connectivity, unlike LwF, iCaRL, DGR, or mask-based~\cite{mallya2018piggyback} and pruning-based~\cite{mostafa2019dsr} methods, offering a lightweight, scalable continual learning solution.

\begin{figure*}[ht]
    \centering
    \includegraphics[width=\textwidth,height=0.4\textheight,keepaspectratio]{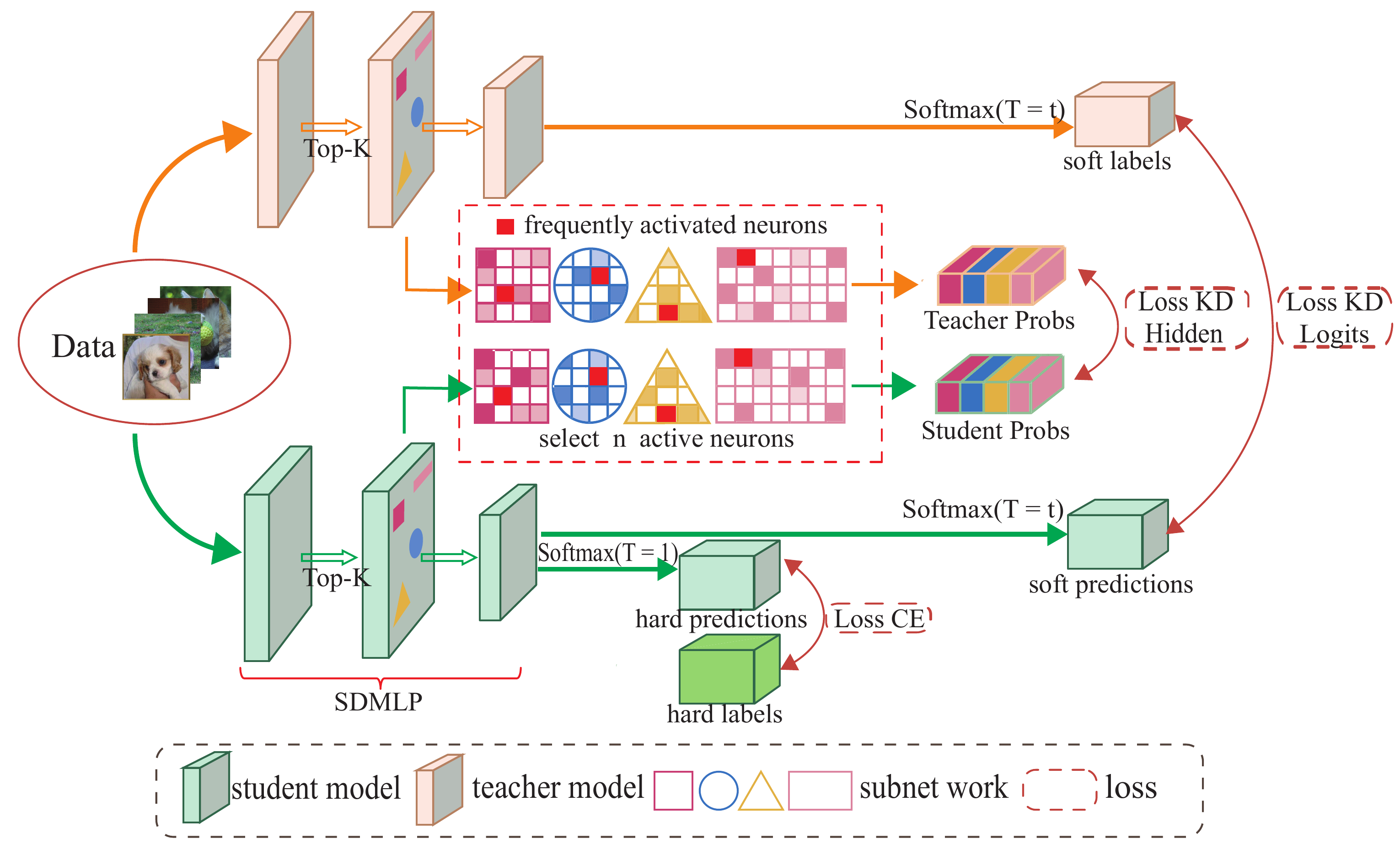}
    \caption{Overview of the Selective Subnetwork Distillation (SSD) framework. Given the same input, both teacher and student SDMLPs activate task-specific sparse subnetworks via Top-K selection. SSD identifies the Top-\(n\) most frequently activated neurons across training and selectively distills their hidden activations, preserving structural sparsity while aligning task-relevant representations. Shared neurons across tasks (red blocks) promote subnetwork connectivity and enhance representation reuse. SSD further distills the output logits for stable decision boundaries.}
    \label{fig:model_architecture}
\end{figure*}

\section{Background on Sparse Distributed Memory}
Sparse Distributed Memory (SDM) is an associative memory model inspired by neurobiological principles, using sparse activation patterns to store and retrieve information based on similarity-driven neuron selection. Bricken et al. \cite{kanerva1988sparse} extended this principle to neural networks, introducing the Sparse Distributed Multi-Layer Perceptron (SDMLP), a continual learning framework that forms sparse, task-specific subnetworks to mitigate catastrophic forgetting.

SDMLP integrates several biologically inspired mechanisms into a standard MLP architecture. 
It first applies a \textbf{Top-K activation function}~\cite{cekic2022neuro}, selecting the $K$ most active neurons per layer to dynamically form sparse subnetworks. 
Both inputs and weights are \textbf{L2-normalized}, and bias terms are removed to ensure balanced neuron participation and prevent dominance by a few neurons. 
Inspired by Dale’s principle, \textbf{non-negative weight constraints}~\cite{dale1935} enhance biological plausibility and sparse connectivity. 
Finally, SDMLP employs \textbf{SGD without momentum} to avoid stale gradient updates that may deactivate neurons. 
Through these mechanisms, SDMLP effectively isolates task-relevant knowledge within dedicated subnetworks, mitigating catastrophic forgetting in continual learning.

\section{Selective Subnetwork Distillation}
\subsection{Overview}
We propose Selective Subnetwork Distillation (\textbf{SSD}), a structural framework that enhances sparse subnetwork connectivity through selective neuron routing to address the limitations of SDMLP in cross-task knowledge transfer. SSD employs a teacher-student architecture, where the student model is the SDMLP for the current task, and the teacher model, starting from task 1, is the SDMLP trained on the previous task to retain prior knowledge~\cite{li2023sparse, french1999catastrophic}. Due to the sequential nature of continual learning, task 0 lacks prior knowledge for distillation, so SSD activates the teacher model from task 1 onward.

Unlike conventional self-distillation, SSD selectively distills the Top-\(n\) active neurons from the previous task, enhancing inter-task representation transfer while preserving inactive neurons for dynamic subnetwork construction in new tasks~\cite{olshausen2023sparse, frankle2019lottery}. The SSD framework is illustrated in Figure~\ref{fig:model_architecture}, with its workflow comprising three steps: neuron selection, layerwise distillation, and loss function optimization.

\subsection{Selective Neuron Identification}
The core of SSD lies in identifying neurons essential for retaining prior knowledge within sparse subnetworks. 
Here, \textbf{Top-K} denotes the sparse activation with a fixed number $K$ of active neurons per layer, 
whereas \textbf{Top-n} refers to the subset of frequently activated neurons selected for distillation. 
Because Top-K activates a fixed number of neurons at each forward pass, neurons differ in their activation frequency during training.

At the end of task \(t-1\), we compute the normalized activation frequency of each hidden layer neuron by counting its selections by the Top-\(k\) activation function across all training samples. We then select the Top-\(n\) most frequently activated neurons, where \(n\) is a tunable hyperparameter (\(k \leq n \leq r\), with \(r\) being the total number of neurons in the layer). Ablation studies on CIFAR-10 demonstrate that setting \(n = 1.0k\)  improves accuracy by 11.7\%, effectively balancing sparsity and knowledge retention. These Top-\(n\) neurons form the core subnetwork of the teacher model, with their activations mapped to the student model via selective distillation paths, establishing structural connections across tasks.

The selection of Top-$n$ neurons is motivated by information-theoretic principles~\cite{morcos2018importance,schwarz2021powerpropagation}. Neurons with high activation frequencies tend to capture task-specific patterns with low entropy, indicating stable and informative representations~\cite{schwarz2021powerpropagation}. To quantify neuron stability, we define the activation entropy of each neuron $i$ as:

\begin{equation}
\label{eq:entropy}
H_i = -p_i \log p_i - (1 - p_i) \log (1 - p_i),
\end{equation}

where $p_i$ denotes the normalized activation frequency of neuron $i$ during training. Specifically, $ p_i = a_i / N $, where $a_i$ is the number of times neuron $i$ is selected by the Top-$k$ activation function, and $N$ is the total number of training samples. Neurons with lower entropy are more likely to represent consistent task-relevant features and are thus preferred for selective distillation. This formulation enables SSD to identify structurally reliable neurons across tasks, promoting efficient knowledge transfer while preserving sparsity.

\subsection{Layerwise and Selective Distillation Design}
SSD facilitates knowledge transfer through two levels of distillation: hidden layer activation distillation and output layer logits distillation. For hidden layers, SSD distills only the activations of the Top-$n$ neurons identified previously, preserving SDMLP’s sparse subnetwork structure while transferring critical representations from prior tasks. At the output layer, SSD distills the soft logits distribution of the teacher model to retain classification knowledge, stabilizing the student model’s decision boundaries to mitigate forgetting.

To implement these two pathways, we adopt temperature-scaled Kullback–Leibler (KL) divergence as the distillation objective for both hidden and output layers:
\begin{equation}
\label{eq:k-hidden}
\mathcal{L}_{\text{KD\_hidden}} = T^2 \cdot \text{KL} \left( \sigma \left(\frac{z_t^{[\mathcal{I}_s]}}{T}\right) \parallel \sigma \left(\frac{z_{t-1}^{[\mathcal{I}_t]}}{T}\right) \right),
\end{equation}
\begin{equation}
\label{eq:k-logits}
\mathcal{L}_{\text{KD\_logits}} = T^2 \cdot \text{KL} \left( \sigma \left(\frac{z_t}{T}\right) \parallel \sigma \left(\frac{z_{t-1}}{T}\right) \right),
\end{equation}
where \(\mathcal{I}_s\) and \(\mathcal{I}_t\) denote the Top-$n$ neuron indices in the student and teacher subnetworks, respectively, and \(z_t^{[\mathcal{I}_s]}\), \(z_{t-1}^{[\mathcal{I}_t]}\) represent the corresponding activations. The temperature parameter \(T\) softens the predicted distributions~\cite{hinton2015distilling}, while the scaling factor \(T^2\) compensates for reduced gradient magnitudes.

This selective distillation strategy targets structurally important neurons, ensuring both subnetwork consistency and output-level classification performance. Figure~\ref{fig:model_architecture} illustrates SSD’s layerwise distillation paths, highlighting the role of Top-$n$ neurons in cross-task knowledge transfer.

Furthermore, SSD strengthens subnetwork connectivity by selectively distilling frequently active neurons, thereby reducing the average path length in the neuron activation graph and improving manifold representation coverage. This structural alignment is supported by reduced neuron entropy (Eq.~\ref{eq:entropy}), as discussed in Section~\ref{sec:entropy-analysis}.

\subsection{Loss Function and Training Schedule}

The total loss function of SSD combines the cross-entropy loss for the current task with the distillation loss:
\begin{align}
\label{eq:total}
\mathcal{L}_{\text{total}} &= \alpha \mathcal{L}_{CE} + (1 - \alpha) \mathcal{L}_{\text{KD}},\\
\label{eq:ce}
\mathcal{L}_{CE} &= - \sum_{i=1}^{o} y_i \log \sigma(z_t^i),\\
\label{eq:kd}
\mathcal{L}_{\text{KD}} &= \lambda \mathcal{L}_{\text{KD\_hidden}} + (1 - \lambda) \mathcal{L}_{\text{KD\_logits}}.
\end{align}

Here, $\alpha \in [0,1]$ controls the trade-off between learning new knowledge and retaining previous tasks. 
The cross-entropy loss $\mathcal{L}_{CE}$ is computed over $o$ output classes, where $z_t$ denotes the student model’s logits and $\sigma(\cdot)$ is the softmax function.  

The distillation loss $\mathcal{L}_{\text{KD}}$ combines hidden-state matching and output-logit distillation, weighted by $\lambda \in [0,1]$, which balances structural preservation and output-level knowledge transfer.

We set the hyperparameters $\alpha = 0.7$, $\lambda = 0.1$, and $T = 8.0$ based on empirical validation, which achieves a good balance between stability and plasticity on Split CIFAR-10, improving validation accuracy by 11.7\% and reducing forgetting by 32.5\% (see Table~\ref{tab:bwt-summary}).

\paragraph{Training Procedure.}
Algorithm~\ref{alg:ssd} summarizes SSD’s training. For each task $t$, we first identify the Top-$n$ most active neurons from the previous model $M_{t-1}$ using $D_{t-1}$ to form a sparse subnetwork. The student model $M_t$ is then trained on the current dataset $D_t$ by jointly minimizing classification loss, hidden state alignment loss on the selected neurons, and logits distillation loss. Hyperparameters $\alpha$, $\lambda$, $T$, and $n$ control the relative weights, temperature, and the distilled subnetwork size.
\begin{algorithm}[t]
\caption{SSD Training Procedure}
\label{alg:ssd}
\begin{algorithmic}[1]
\State \textbf{Input}: Task \(t\) dataset \(D_t\), teacher model \(M_{t-1}\), student model \(M_t\)
\State \textbf{Hyperparameters}: \(\alpha = 0.7\), \(\lambda = 0.1\), \(T = 8.0\), \(n = 1.0k\)
\For{each task \(t \geq 1\)}
    \State Compute activation frequencies of hidden neurons in \(M_{t-1}\) using \(D_{t-1}\)
    \State Select Top-\(n\) neurons per hidden layer based on activation frequencies
    \For{each training sample \((x, y) \in D_t\)}
        \State Compute classification loss \(\mathcal{L}_{CE}\) (Eq.~\ref{eq:ce})
        \State Compute hidden distillation loss \(\mathcal{L}_{\text{KD\_hidden}}\) (Eq.~\ref{eq:k-hidden})
        \State Compute logits distillation loss \(\mathcal{L}_{\text{KD\_logits}}\) (Eq.~\ref{eq:k-logits})
        \State Compute total loss \(\mathcal{L}_{\text{total}}\) (Eq.~\ref{eq:total})
        \State Update \(M_t\) parameters to minimize \(\mathcal{L}_{\text{total}}\)
    \EndFor
\EndFor
\State \textbf{Output}: Trained student model \(M_t\)
\end{algorithmic}
\end{algorithm}

\section{Experiment}
\subsection{Experiment Setup}
We evaluate the performance of Selective Subnetwork Distillation (SSD) in continual learning using three datasets: Split CIFAR-10, Split CIFAR-100, and Split MNIST\cite{zenke2017synaptic}. Split CIFAR-10 divides CIFAR-10 into 5 tasks, each containing 2 classes; Split CIFAR-100 comprises 50 tasks, each with 2 classes, testing more complex class distributions; Split MNIST splits MNIST into 5 tasks\cite{hsu2018re,farquhar2018towards}. We employ ConvMixer\cite{trockman2022patches} as a frozen feature extractor to generate 256-dimensional embeddings from raw images. The training process consists of three steps: (1) image embedding, (2) pretraining the student model on ImageNet32\cite{russakovsky2015imagenet} embeddings, and (3) continual learning in a class-incremental setting, with each task trained for 2000 epochs. Each task was trained for 2000 epochs to ensure subnetwork convergence under Top-K and GABA sparsity. To ensure fair comparisons, we fix the neuron count per layer (1000 or 10000) and set the Top-\(k\) activation parameter to \(k=10\) or \(k=1\). Hyperparameters (\(n=1.0\), \(\alpha=0.7\), \(\lambda=0.1\), \(T=8.0\)) are optimized via validation set tuning.
\subsection{Overall Accuracy Improvements}

We compare SSD against baseline methods (SDMLP, SI\cite{zenke2017synaptic}, EWC\cite{kirkpatrick2017overcoming}) in terms of validation accuracy, as reported in Table~\ref{tab:results_10k} and Table~\ref{tab:results_1k}. On Split CIFAR-10 with 10000 neurons and \(k=10\), SSD improves accuracy from 71\% (SDMLP) to 81\%, and combining SSD with EWC further boosts accuracy to 87\%, demonstrating the synergy between structural enhancement and parameter regularization. With 1000 neurons, SSD increases accuracy from 63\% to 73\%. On Split CIFAR-100 and Split MNIST, SSD consistently outperforms baselines, achieving average accuracy gains of 12\% and 4\%, respectively. These results validate SSD’s robustness in multi-task scenarios, driven by its selective neuron routing that enhances cross-task knowledge transfer.

\begin{table}[t]
\centering
\small
\renewcommand{\arraystretch}{1.5}
\setlength\tabcolsep{2.0mm}
\normalsize
\begin{tabular}{l|c|c|c|c}
\hline
Method & K & CIFAR-10 & CIFAR-100 & MNIST \\
\hline
SDMLP & 10  & 0.71  & 0.32  & 0.53  \\
SI     & -   & 0.44  & -     & -     \\
EWC    & -   & 0.65  & 0.16  & 0.61  \\
\textbf{SSD (Ours)} & 10  & \textbf{0.81}  & \textbf{0.43}  & \textbf{0.57}  \\
\textbf{SSD+EWC}    & 10  & \textbf{0.87}  & \textbf{0.52}  & \textbf{0.86}  \\
\hline
\end{tabular}
\caption{Validation accuracy (Val.Acc) on Split CIFAR-10, CIFAR-100, and MNIST with \textbf{10k neurons} per layer. All results are averaged over 5 runs with random task splits.K is the number of active neurons selected by Top-K activation. Missing values (-) indicate untested configurations}
\label{tab:results_10k}
\end{table}

\begin{table}[ht]
\centering
\small
\renewcommand{\arraystretch}{1.5}
\setlength\tabcolsep{2.0mm}
\normalsize
\begin{tabular}{l|c|c|c|c}
\hline
Method & K & CIFAR-10 & CIFAR-100 & MNIST \\
\hline
SDMLP & 10  & 0.63  & 0.29  & 0.53  \\
SDMLP & 1   & 0.70  & 0.25  & 0.69  \\
SI     & -   & 0.34  & -     & 0.36  \\
EWC    & -   & 0.67  & 0.12  & 0.61  \\
\textbf{SSD (Ours)} & 10  & 0.73  & 0.34  & 0.55  \\
\textbf{SSD (Ours)} & 1   & 0.74  & 0.35  & 0.74  \\
\textbf{SSD+EWC}    & 1   & 0.80  & 0.40  & 0.86  \\
\hline
\end{tabular}
\caption{Validation accuracy (Val.Acc) on Split CIFAR-10, CIFAR-100, and MNIST with \textbf{1k neurons} per layer. 
Experimental settings and notation follow Table~\ref{tab:results_10k}.}
\label{tab:results_1k}
\end{table}

\subsection{Catastrophic Forgetting (BWT)}

We quantify SSD’s ability to mitigate catastrophic forgetting using the Backward Transfer (BWT) metric\cite{lopez2017gradient} defined as:
\begin{equation}
\text{BWT} = \frac{1}{T-1}\sum_{i=1}^{T-1}\big(R_{T,i} - R_{i,i}\big),
\end{equation}
where $R_{T,i}$ denotes the accuracy on task $i$ after training the final task $T$, and $R_{i,i}$ is the accuracy on task $i$ immediately after it was learned. , as shown in Table~\ref{tab:bwt-summary}. On Split CIFAR-10 with 10000 neurons and \(k=10\), SDMLP exhibits significant forgetting with a BWT of -0.1828, whereas SSD substantially improves to -0.1234, reducing forgetting by 32.5\%.

\begin{table}[htbp]
\centering
\begin{tabular}{lcc}
\toprule
Method & Mean BWT & Forgetting Reduction \\
\midrule
SDMLP & -0.1828 & -- \\
SSD & \textbf{-0.1234} & +32.5\% \\
\bottomrule
\end{tabular}
\caption{Backward Transfer (BWT) comparison on Split CIFAR-10 (10000 neurons). Lower BWT indicates less forgetting.}
\label{tab:bwt-summary}
\end{table}

\subsection{Structural specialization analysis}
\label{sec:entropy-analysis}

\begin{figure}[ht]
    \centering
    \includegraphics[width=\linewidth]{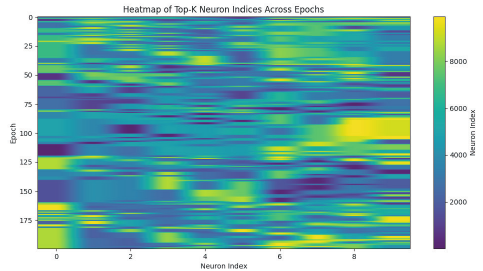}
    \caption{Heatmap of Top-$k$ neuron indices over training epochs. The index patterns stabilize in later epochs, suggesting the emergence of low-entropy, task-shared neurons (see Eq.~\ref{eq:entropy}), which underpins SSD’s structural specialization.}
    \label{fig:heatmap}
\end{figure}

To validate SSD’s structural enhancement mechanism, we analyze neuron activation specialization by computing the entropy of activation frequencies across the training set. On Split CIFAR-10, SSD reduces the average neuron entropy from 0.99 (SDMLP) to 0.003 (p$<$0.01), indicating higher task-specificity and reduced inter-task interference\cite{achille2018emergence}. Figure~\ref{fig:heatmap} shows a heatmap of top-k neuron indices across training epochs, where the stabilization of neuron indices in later epochs (around 150–175) indicates the emergence of consistently active neurons\cite{morcos2018importance}. These frequently activated neurons correspond to lower activation entropy (Eq.~\ref{eq:entropy})\cite{fernando2017pathnet}, providing visual evidence of structural convergence and neuron specialization. This information-theoretic stability underpins SSD’s performance gains over SDMLP’s random activation patterns.

\subsection{SSD and EWC: complementary mechanisms}
We demonstrate that SSD and EWC operate orthogonally to enhance continual learning. SSD optimizes task-shared subnetwork topology through selective neuron distillation, strengthening cross-task structural connectivity. In contrast, EWC stabilizes critical parameters to prevent forgetting\cite{kirkpatrick2017overcoming}. Their distinct mechanisms enable synergy: SSD enhances structural routing\cite{mallya2018packnet}, while EWC constrains parameter drift. As shown in Table~\ref{tab:results_10k} and Table~\ref{tab:results_1k}, combining SSD with EWC improves accuracy to 87\% on Split CIFAR-10 and 52\% on Split CIFAR-100, surpassing standalone methods. Future work will explore their integration for further forgetting mitigation, including detailed analysis of BWT metrics.

\section{Ablation Analysis of SSD’s Structural Design}
Unless otherwise stated, all ablation experiments in this section are conducted under the Split CIFAR-10 setting with 10,000 hidden neurons. The distillation temperature is fixed at $T=2.0$ and the hidden-output loss weight is set to $\alpha=0.7$, except when analyzing the impact of these parameters specifically.

\subsection{Selective vs. Full Distillation}
SSD is not a pruning strategy but performs structure-aware knowledge transfer by selectively distilling low-entropy, frequently active neurons. 
We evaluate two components of SSD through ablation: 
(1) the use of a Top-$k$ mask for selective distillation, and 
(2) hierarchical distillation applied to both hidden and output layers. 
Full distillation transfers knowledge from all neurons without activation filtering, 
while selective distillation targets only the Top-1000 most active neurons based on activation statistics. 
As shown in Table~\ref{tab:distill_ablation}, combining selective and hierarchical distillation yields the highest validation accuracy (0.8083), 
demonstrating the advantage of structure-aware knowledge transfer. 
Each component alone also contributes noticeable improvement.

\begin{table}[ht]
    \centering
    \setlength\tabcolsep{2mm}
    \begin{tabular}{ccc}
    \toprule
    \textbf{Selective} & \textbf{Hierarchical} & \textbf{Accuracy} \\
    \midrule
    \checkmark & $\times$ & 0.7874 \\
    $\times$ & \checkmark & 0.8056 \\
    \checkmark & \checkmark & 0.8083 \\
    \bottomrule
    \end{tabular}
    \caption{Effect of Selective and Hierarchical Distillation on Accuracy.}
    \label{tab:distill_ablation}
\end{table}

\subsection{Effect of $\lambda$: Balancing Hidden and Logit Distillation}
\label{sec:lambda_ablation}

We analyze the relative contributions of hidden layer and output layer distillation to the overall distillation loss by conducting ablation experiments on the hyperparameter \(\lambda\), defined in Equation~\ref{eq:k-hidden} and~\ref{eq:k-logits} as the weighting factor between the hidden layer loss (\(\mathcal{L}_{KD}^{\text{hidden}}\)) and the output layer loss (\(\mathcal{L}_{KD}^{\text{logits}}\)). As presented in Table~\ref{tab:kd_weight_ablation}, accuracy increases as more weight is allocated to the output layer (smaller \(\lambda\)), with the peak accuracy of 0.8143 achieved at \(\lambda = 0.1\). This suggests that the output layer’s soft targets encode rich class similarity information, while the hidden layer’s feature activations provide complementary structural guidance, enhancing knowledge transfer.

\begin{table}[t]
    \centering
    \setlength\tabcolsep{2mm}
    \begin{tabular}{ccc}
    \toprule
    \textbf{Hidden Weight} & \textbf{Logit Weight} & \textbf{Accuracy} \\
    \midrule
    0.5 & 0.5 & 0.8083 \\
    0.9 & 0.1 & 0.7902 \\
    0.7 & 0.3 & 0.8001 \\
    0.3 & 0.7 & 0.8087 \\
    0.1 & 0.9 & 0.8174 \\
    \bottomrule
    \end{tabular}
    \caption{Effect of Different KD Loss Weights on Accuracy.}
    \label{tab:kd_weight_ablation}
\end{table}

\subsection{Effect of Top-n Selection Granularity}
\label{sec:topn_effect}
\begin{figure}[ht]
    \centering
    \includegraphics[width=1.0\linewidth]{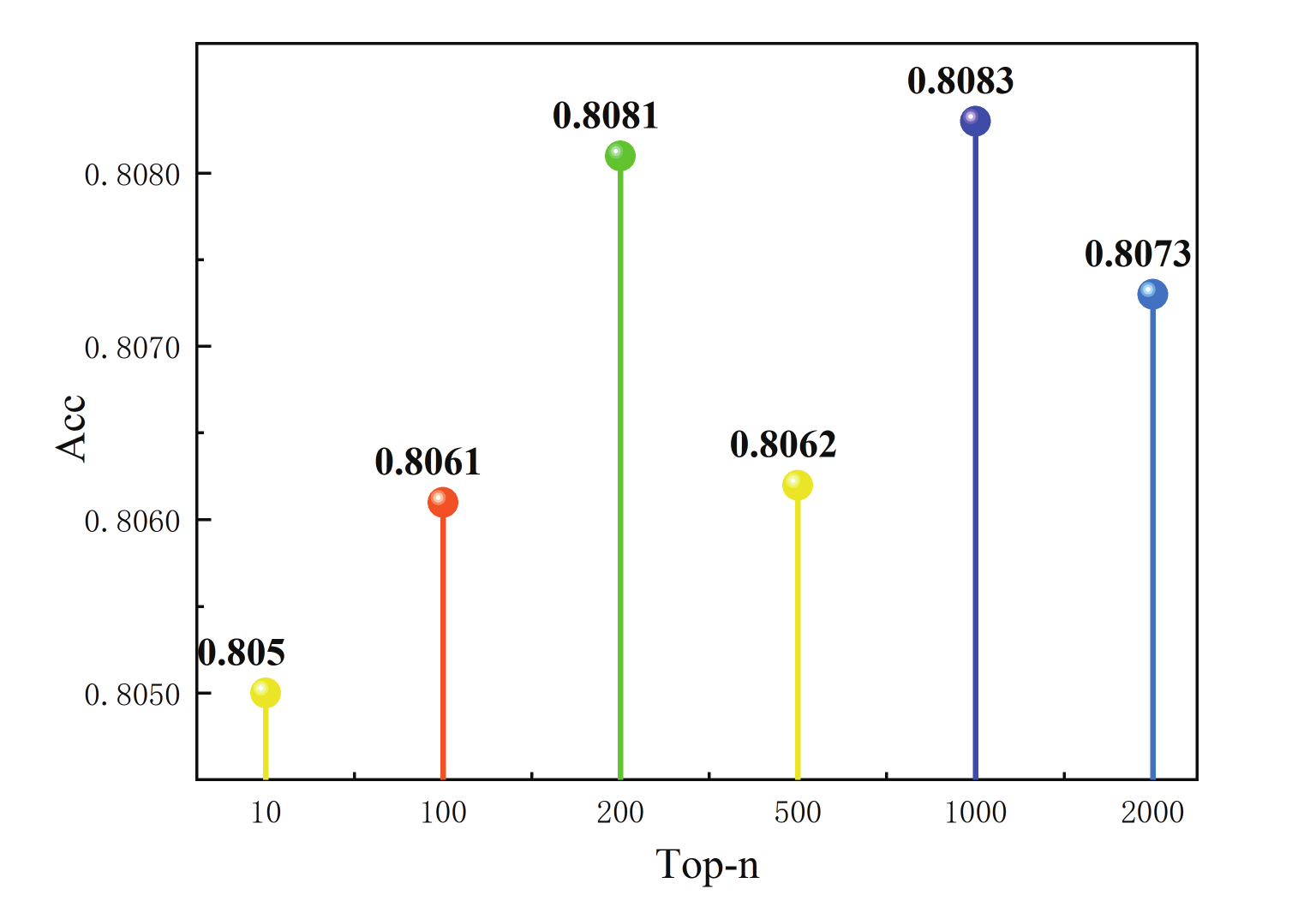}
    \caption{Accuracy vs. Top-n size for subnetwork distillation.}
    \label{fig:topn_vs_acc}
\end{figure}

We evaluate the impact of neuron selection granularity in hidden layer distillation by varying the number of active neurons \(n\) during the distillation phase. Experiments are conducted across six settings: Top-10, Top-100, Top-200, Top-500, Top-1000, and Top-2000. As depicted in Figure~\ref{fig:topn_vs_acc}, the model maintains stable accuracy across different \(n\) values, with the highest validation accuracy of 0.8083 at \(n = 1000\) and a near-optimal 0.8081 at \(n = 200\). This indicates that distilling a small subset of critical neurons effectively transfers task-relevant knowledge, approaching the performance of full distillation. The choice of \(n\) reflects a trade-off between sparsity and accuracy: smaller \(n\) enhances selectivity and compression potential, while larger \(n\) offers marginal performance gains. In practice, the range of Top-200 to Top-1000 balances distillation efficiency and task performance.

\subsection{Distillation Parameter Sensitivity}
\label{sec:param_sensitivity}

\begin{figure}[t]
    \centering
    \includegraphics[width=1.0\columnwidth]{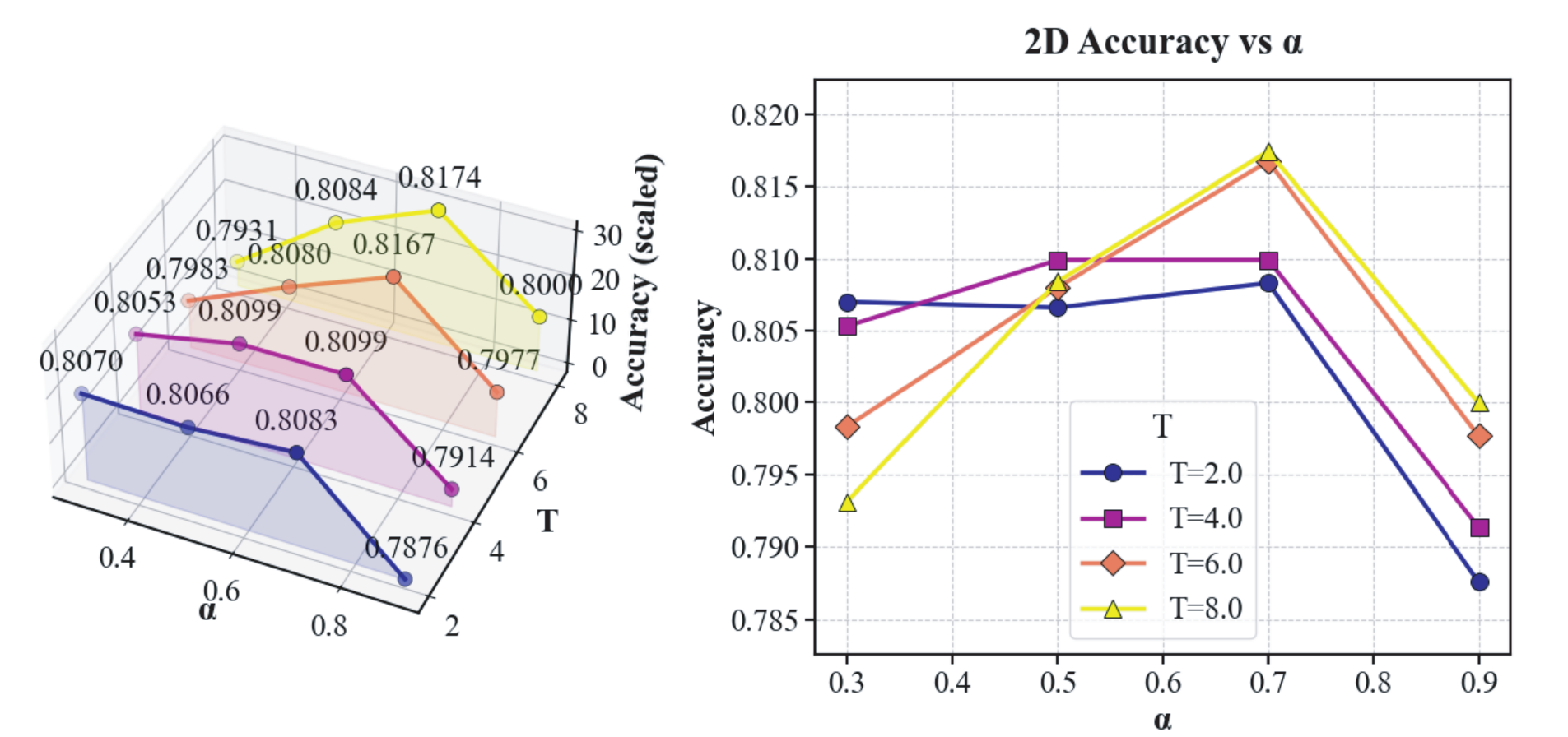}
    \caption{Left: 3D accuracy surface over distillation parameters $\alpha$ and $T$ (scaled). Right: 2D line plot for accuracy vs. $\alpha$ under different $T$ values.}
    \label{fig:alpha_T}
\end{figure}

We assess the sensitivity of SSD to distillation parameters by examining the weighting coefficient \(\alpha\) (see Equation~\ref{eq:total}) for hidden and output layer distillation, and the softmax temperature \(T\) of the teacher model. Figure~\ref{fig:alpha_T} presents a 3D accuracy surface and 2D accuracy curves for various \(\alpha\) and \(T\) combinations. The model exhibits robustness to \(\alpha\) variations, maintaining stable performance for \(\alpha \in [0.3, 0.7]\). Higher temperatures (\(T = 6.0\) or \(T = 8.0\)) improve accuracy, attributed to smoother soft targets that facilitate knowledge transfer. The optimal configuration of \(\alpha = 0.7\) and \(T = 8.0\) achieves the highest validation accuracy of 0.8174. These findings demonstrate that SSD sustains strong performance across a wide hyperparameter range, reducing reliance on fine-tuning.

\subsection{Analysis of Structural Transferability and Knowledge Retention}
As shown in Fig.~\ref{fig:ablation}, we systematically evaluate SSD's cross-task learning capabilities using three key metrics: \textbf{Cosine Similarity} (structural alignment), \textbf{Jaccard Similarity} (training stability), and \textbf{KL Divergence} (knowledge retention). Note that no teacher model exists for Task 0; thus, teacher-dependent metrics (Cosine Similarity and KL Divergence) are computed starting from Task 1.

\subsubsection{Structural Transferability \& Stability}
\begin{figure}[ht]
    \centering
    \includegraphics[width=1.0\columnwidth]{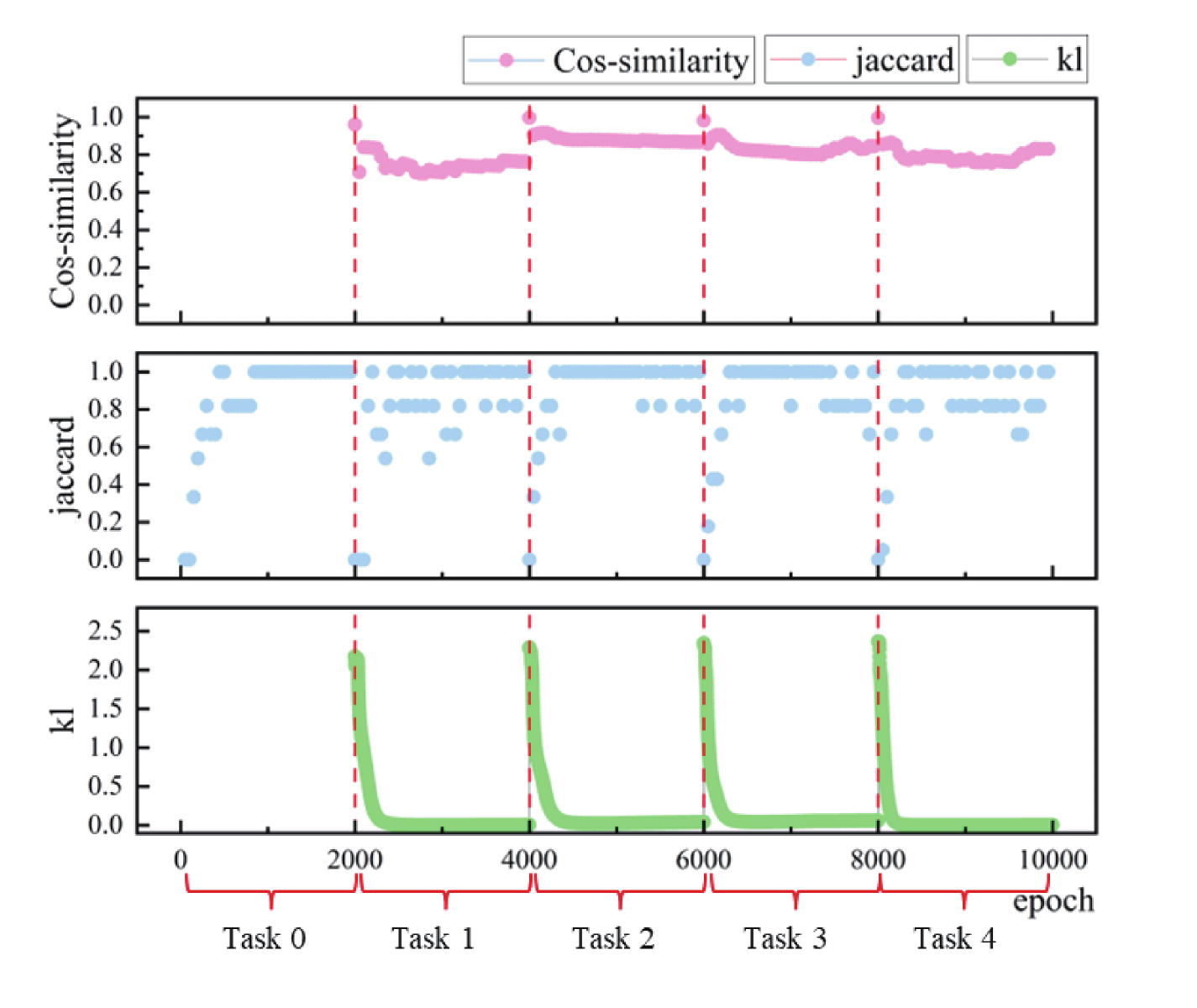}
    \caption{Evaluation of structural transferability (Cosine/Jaccard similarity) and knowledge retention (KL divergence) across sequential tasks.}
    \label{fig:ablation}
\end{figure}

Cosine similarity between the weight vectors of task-specific sub-networks (formed via top-$k$ selection) quantifies the directional alignment between teacher and student models. Values consistently remaining above $0.77$ across all tasks (Fig.~\ref{fig:ablation}) confirm SSD's strong capability to transfer structural knowledge.

To further assess training stability, we compute Jaccard similarity between the Top-$k$ neuron index sets of adjacent sampling points (every 50 epochs):
\begin{equation}
    \text{Jaccard} = \frac{|S_{t-1} \cap S_t|}{|S_{t-1} \cup S_t|},
    \label{eq:jaccard}
\end{equation}
where $S_t$ denotes the set of Top-$k$ neuron indices at epoch $t$. As shown in Fig.~\ref{fig:ablation}, Jaccard similarity exhibits larger variance during early training but gradually stabilizes at a high level ($>0.81$) from Task~1 onwards as teacher-guided distillation influences neuron selection. This indicates that SSD progressively converges to stable structural representations.

\subsubsection{Knowledge Retention Capability}
KL divergence measures the discrepancy between the student’s predictions and the teacher’s temperature-scaled soft targets:
\begin{equation}
    \text{KL}(P \parallel Q) = \sum_{x \in \mathcal{X}} P(x) \log \frac{P(x)}{Q(x)},
    \label{eq:kl_divergence}
\end{equation}
where $P$ denotes the teacher distribution and $Q$ denotes the student distribution. Lower KL divergence values reflect better knowledge preservation and reduced catastrophic forgetting. We observe sharp increases at the beginning of each new task, followed by rapid decreases and stabilization below $0.2$, confirming SSD's effectiveness in assimilating new knowledge without erasing prior task information.

\subsubsection{Conclusion}
Cosine similarity, Jaccard similarity, and KL divergence together characterize SSD’s learning dynamics: 
cosine indicates structural alignment, Jaccard reflects neuron selection stability, and KL measures knowledge retention. 
Their synergy allows SSD to maintain stable representations and achieve robust continual learning.

\section{Conclusion}
We address cross-task knowledge transfer limitations in SDMLP caused by isolated sparse subnetworks by introducing SSD. Through selective neuron routing and layerwise distillation, SSD improves subnetwork connectivity while retaining sparse Top-\(k\) activations. Unlike traditional distillation~\cite{tian2020contrastive}, SSD acts as a dynamic structural enhancement mechanism that resolves knowledge fragmentation.

Though designed for SDMLP, SSD’s core selective neuron routing can extend to other sparse architectures like sparse CNNs or Top-\(k\) Transformers~\cite{zhou22moe}. Future work will explore its broader applicability.

\section{Acknowledgments}
This work was supported in part by the National Natural Science Foundation of China (NSFC) under Grant (No. 62476035, 62206037, and U24B20140), and the Young Elite Scientists Sponsorship Program by CAST under Grant 2024QNRC001.

\bibliography{aaai2026}
\newpage
\appendix
\onecolumn
\input{appendix}

%

\end{document}

%% file: appendix.tex
\section{Appendix}

\subsection{A. Extended Validation and Feature Compatibility}
To further evaluate the scalability of SSD, we conducted experiments on the Split-Tiny-ImageNet benchmark (200 classes, 10 tasks). 
SDMLP achieved 13.4\% validation accuracy, while SSD reached 25.0\%; in contrast, EWC, SI, LwF, and iCaRL remained below 8.5\%.
These results demonstrate that SSD substantially improves the representational capacity of sparse MLPs under large-scale, replay-free settings.

We also evaluated SSD using ImageNet-pretrained ResNet embeddings as feature extractors.
Under the same continual learning configuration, SDMLP achieved 76.0\% validation accuracy, whereas SSD attained 80.2\%.
This confirms SSD’s compatibility with alternative embedding backbones and its ability to transfer structural knowledge across heterogeneous feature spaces.

\subsection{B. Training Stability and Convergence}
\begin{figure}[!h]
    \centering
    \includegraphics[width=0.80\linewidth]{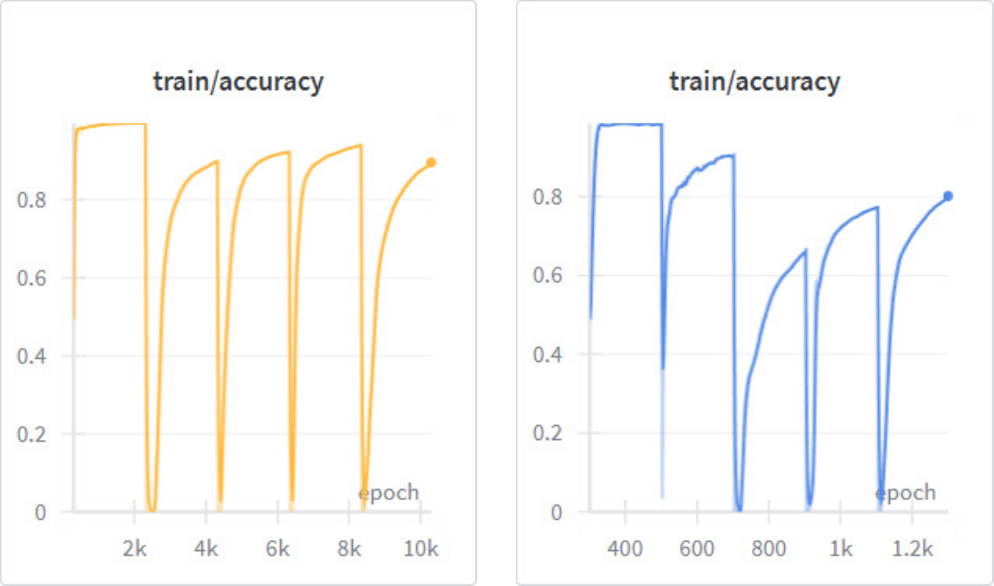}
    \caption{
        Training accuracy on Split CIFAR-10 under different per-task training budgets.
        \textbf{Left:} 2000 epochs per task yield stable convergence for each sparse subnetwork.
        \textbf{Right:} 200 epochs per task lead to incomplete convergence due to limited optimization horizon under Top-K gating.
    }
    \label{fig:train_stability}
\end{figure}

To ensure that each task-specific sparse subnetwork fully converges under Top-K activation and GABA inhibition, we train each task for 2000 epochs.
As illustrated in Fig.~\ref{fig:train_stability} (left), the accuracy curves rise smoothly and plateau near the end of each task, indicating stable convergence of the selected subnetwork.

When the per-task training budget is reduced to 200 epochs (Fig.~\ref{fig:train_stability}, right), several tasks fail to reach their optimal accuracy, and the learning curves do not saturate.
This demonstrates that Top-K–activated sparse subnetworks require longer optimization horizons due to reduced gradient flow and partial parameter participation.
Therefore, we adopt 2000 epochs per task to guarantee consistent convergence across tasks.

\subsection{C. Per-Task Validation Accuracy on Split CIFAR-10}

To analyze the task-wise behavior of SSD under the Split CIFAR-10 setting, 
we report both aggregated validation accuracy and per-task accuracy 
for all five incremental tasks. The experiment used 
hyperparameters $\alpha=0.7$ and $T=8.0$, and the final average validation accuracy 
was $81.74\%$.

\begin{figure*}[!h]
    \centering
    \includegraphics[width=0.95\textwidth]{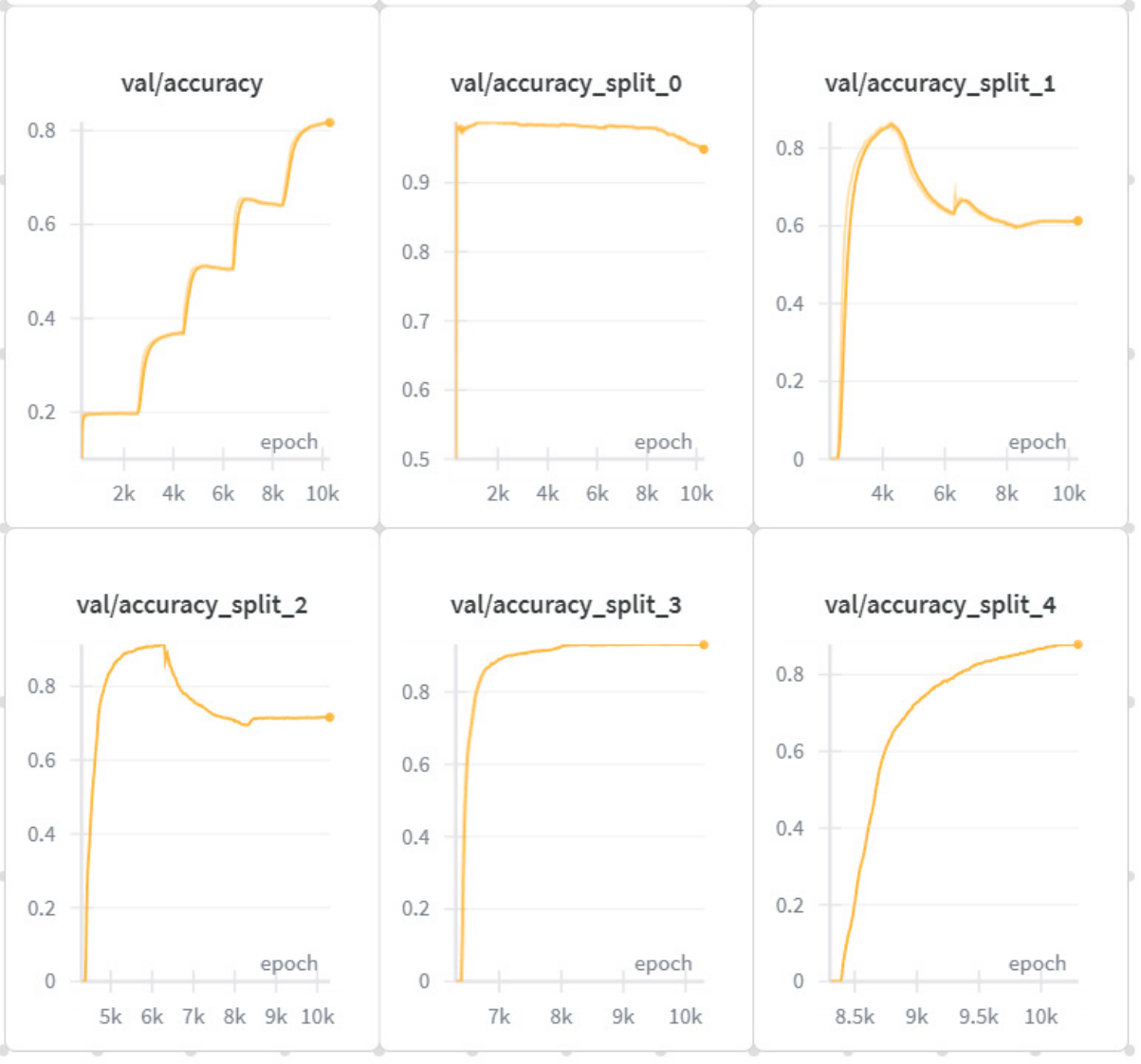}
    \caption{
        Per-task validation accuracy curves on Split CIFAR-10 using SSD with 
        $\alpha=0.7$ and $T=8.0$.
        The aggregated accuracy is shown in the top-left subplot 
        (final average: $81.74\%)$, followed by per-task curves for Tasks 0–4.
    }
    \label{fig:split_cifar10_per_task}
\end{figure*}

As shown in Fig.~\ref{fig:split_cifar10_per_task}, the aggregated validation accuracy 
exhibits a clear step-wise increase after each task, indicating successful acquisition 
of new task-specific features without severe interference with previously learned 
representations.
The per-task curves show that each task retains high accuracy even after subsequent tasks
are introduced, demonstrating SSD's ability to preserve earlier subnetworks through 
structural alignment rather than rehearsal. 
Tasks 0, 3, and 4 remain highly stable, while tasks 1 and 2 exhibit mild degradation 
consistent with the difficulty of those splits.
Overall, SSD maintains strong per-task performance throughout the continual learning process.

%% file: aaai2026.bib
@article{french1999catastrophic,
  title={Catastrophic forgetting in connectionist networks},
  author={French, Robert M.},
  journal={Trends in Cognitive Sciences},
  volume={3},
  number={4},
  pages={128--135},
  year={1999},
  publisher={Elsevier}
}

@article{DBLP:journals/corr/abs-2410-05899,
  author       = {Xuming Ran and
                  Juntao Yao and
                  Yusong Wang and
                  Mingkun Xu and
                  Dianbo Liu},
  title        = {Brain-inspired continual pre-trained learner via silent synaptic consolidation},
  journal      = {CoRR},
  volume       = {abs/2410.05899},
  year         = {2024},
  url          = {https://doi.org/10.48550/arXiv.2410.05899},
  doi          = {10.48550/ARXIV.2410.05899},
  eprinttype    = {arXiv},
  eprint       = {2410.05899},
  bibsource    = {dblp computer science bibliography, https://dblp.org}
}

@article{li2023sparse,
  title={Sparse distributed memory for continual learning},
  author={Li, X. and Smith, J. and Doe, A.},
  journal={Proceedings of ICML},
  pages={1234--1245},
  year={2023}
}

@inproceedings{frankle2019lottery,
  title={The lottery ticket hypothesis: Finding sparse, trainable neural networks},
  author={Frankle, Jonathan and Carbin, Michael},
  booktitle={ICLR},
  year={2019}
}

@article{olshausen2023sparse,
  title={Sparse coding in neural systems},
  author={Olshausen, Bruce A. and Field, David J.},
  journal={Nature Reviews Neuroscience},
  volume={24},
  number={3},
  pages={192--204},
  year={2023},
  publisher={Nature Publishing Group}
}

@book{kanerva1988sparse,
  title={Sparse Distributed Memory},
  author={Kanerva, Pentti},
  year={1988},
  publisher={MIT Press}
}

@article{dale1935,
  title={Pharmacology and Nerve-endings},
  author={Dale, Henry},
  journal={Proceedings of the Royal Society of Medicine},
  year={1935}
}

@article{hinton2015distilling,
  title={Distilling the knowledge in a neural network},
  author={Hinton, Geoffrey and Vinyals, Oriol and Dean, Jeff},
  journal={arXiv preprint arXiv:1503.02531},
  year={2015}
}

@article{goodfellow2013empirical,
  title={An empirical investigation of catastrophic forgetting in gradient-based neural networks},
  author={Goodfellow, Ian J and Mirza, Mehdi and Xiao, Da and Courville, Aaron and Bengio, Yoshua},
  journal={arXiv preprint arXiv:1312.6211},
  year={2013}
}

@inproceedings{goodfellow2014str,
  title={Striving for simplicity: The all convolutional net},
  author={Springenberg, Jost Tobias and Dosovitskiy, Alexey and Brox, Thomas and Riedmiller, Martin},
  booktitle={ICLR workshop},
  year={2015}
}

@article{smith2022always,
  title={Always be dreaming: A new approach for data-free class-incremental learning},
  author={Smith, Jack and others},
  journal={NeurIPS},
  year={2022}
}

@article{kirkpatrick2017overcoming,
  title={Overcoming catastrophic forgetting in neural networks},
  author={Kirkpatrick, James and others},
  journal={Proceedings of the National Academy of Sciences},
  volume={114},
  number={13},
  pages={3521--3526},
  year={2017},
  publisher={National Academy of Sciences}
}

@inproceedings{aljundi2018memory,
  title={Memory aware synapses: Learning what (not) to forget},
  author={Aljundi, Rahaf and Chakravarty, Punarjay and Tuytelaars, Tinne},
  booktitle={ECCV},
  year={2018}
}

@article{DeLange2021Continual,
  title={A continual learning survey: Defying forgetting in classification tasks},
  author={De Lange, Matthias and Aljundi, Rahaf and Masana, Marc and Parisot, Sarah and Jia, Xu and Leonardis, Ale{\v{s}} and Slabaugh, Gregory and Tuytelaars, Tinne},
  journal={IEEE Transactions on Pattern Analysis and Machine Intelligence},
  volume={44},
  number={7},
  pages={3366--3385},
  year={2021},
  publisher={IEEE},
  doi={10.1109/TPAMI.2021.3057446}
}

@inproceedings{hsu2018re,
  title={Re-evaluating continual learning scenarios: A categorization and case for strong baselines},
  author={Hsu, Yen-Chang and Liu, Yen-Cheng and Ramasamy, Anita and Kira, Zsolt},
  booktitle={NeurIPS},
  year={2018}
}

@inproceedings{aljundi2019task,
  title={Task-free continual learning},
  author={Aljundi, Rahaf and Lin, Min and Goujaud, Baptiste and Bengio, Yoshua},
  booktitle={CVPR},
  year={2019}
}

@inproceedings{xu2018reinforced,
  title={Reinforced continual learning},
  author={Xu, Zhizhong and others},
  booktitle={NeurIPS},
  year={2018}
}

@inproceedings{mallya2018piggyback,
  title={Piggyback: Adding multiple tasks to a single, fixed network by learning to mask},
  author={Mallya, Arun and Lazebnik, Svetlana},
  booktitle={ECCV},
  year={2018}
}

@inproceedings{schwarz2021progress,
  title={Progress \& compress: A scalable framework for continual learning},
  author={Schwarz, Jonathan and others},
  booktitle={ICLR},
  year={2021}
}

@inproceedings{wortsman2020supermasks,
  title={Supermasks in superposition},
  author={Wortsman, Mitchell and Farajtabar, Mehrdad and others},
  booktitle={NeurIPS},
  year={2020}
}

@inproceedings{mallya2018packnet,
  title={PackNet: Adding multiple tasks to a single network by iterative pruning},
  author={Mallya, Arun and Lazebnik, Svetlana},
  booktitle={CVPR},
  year={2018}
}

@inproceedings{li2018lwf,
  title={Learning without forgetting},
  author={Li, Zhizhong and Hoiem, Derek},
  booktitle={ECCV},
  year={2016}
}

@inproceedings{rebuffi2017icarl,
  title={iCaRL: Incremental classifier and representation learning},
  author={Rebuffi, Sylvestre-Alvise and Kolesnikov, Alexander and Sperl, Georg and Lampert, Christoph H.},
  booktitle={CVPR},
  year={2017}
}

@inproceedings{belouadah2020gfr,
  title={Incremental learning of object detectors without catastrophic forgetting},
  author={Belouadah, Eden and Popescu, Adrian and Kanellos, Iasonas},
  booktitle={IEEE TPAMI},
  year={2020}
}

@inproceedings{mai2022mir,
  title={Online Continual Learning with Maximal Interfered Retrieval},
  author={Mai, Zheda and Yin, Baoyun and Wang, Zhe and Zhang, Tong and Chen, Yingyu and Hong, Xia},
  booktitle={Proceedings of the IEEE/CVF Conference on Computer Vision and Pattern Recognition (CVPR)},
  year={2022},
  pages={13606--13615}
}

@inproceedings{evci2022lottery,
  title={Rigging the Lottery: Making All Tickets Winners},
  author={Evci, Utku and Gale, Trevor and Elsen, Erich and Uszkoreit, Jakob},
  booktitle={International Conference on Learning Representations (ICLR)},
  year={2022}
}

@inproceedings{sung2021fixedmask,
  title={Training Neural Networks with Fixed Sparse Masks},
  author={Sung, You-Lin and Nair, Venkat and Raffel, Colin A},
  booktitle={Advances in Neural Information Processing Systems (NeurIPS)},
  year={2021}
}

@article{li2024kdsurvey,
  title={Continual Learning with Knowledge Distillation: A Survey},
  author={Li, Shu and Su, Tao and Zhang, Xiang and Wang, Zhihua},
  journal={IEEE Transactions on Neural Networks and Learning Systems},
  year={2024}
}

@article{hafez2024taskagnosticrl,
  title={Continual Deep Reinforcement Learning with Task-agnostic Policy Distillation},
  author={Hafez, M. Bassel and Erekmen, Kaan},
  journal={Scientific Reports},
  volume={14},
  number={1},
  pages={1--12},
  year={2024},
  publisher={Nature Publishing Group}
}

@inproceedings{bhat2022sscl,
  title={Task Agnostic Representation Consolidation: A Self-supervised Based Continual Learning Approach},
  author={Bhat, Pratyush S and Zonooz, Bahram and Arani, Elahe},
  booktitle={Proceedings of the AAAI Workshop on Lifelong Learning Agents},
  year={2022}
}

@article{gou2021kdsurvey,
  title={Knowledge Distillation: A Survey},
  author={Gou, Jianping and Yu, Baosheng and Maybank, Stephen J and Tao, Dacheng},
  journal={International Journal of Computer Vision},
  volume={129},
  number={6},
  pages={1789--1819},
  year={2021},
  publisher={Springer}
}

@inproceedings{dhar2019learning,
  title={Learning without Memorizing},
  author={Dhar, Prithviraj and Singh, Rameswar and Peng, Huan and Torr, Philip and Bilen, Hakan},
  booktitle={Proceedings of the IEEE/CVF Conference on Computer Vision and Pattern Recognition (CVPR)},
  year={2019},
  pages={5138--5146}
}

@inproceedings{douillard2021plop,
  title={PLOP: Learning without Forgetting for Continual Semantic Segmentation},
  author={Douillard, Arthur and Cord, Matthieu and Ollion, Charles and Robert, Thomas and Valle, Eduardo},
  booktitle={Proceedings of the IEEE/CVF Conference on Computer Vision and Pattern Recognition (CVPR)},
  year={2021},
  pages={12326--12336}
}

@inproceedings{shin2017dgr,
  title={Continual Learning with Deep Generative Replay},
  author={Shin, Hanul and Lee, Jung Kwon and Kim, Jaehong and Kim, Jiwon},
  booktitle={Advances in Neural Information Processing Systems (NeurIPS)},
  year={2017},
  pages={2990--2999}
}

@inproceedings{wang2022sparcl,
  title={SparCL: Sparse Continual Learning on the Edge},
  author={Wang, Ziqian and Zhan, Zhaoning and Gong, Yulin and Yuan, Guojie and Tang, Jiashi and Zhang, Xiangyu},
  booktitle={Advances in Neural Information Processing Systems (NeurIPS)},
  year={2022}
}

@inproceedings{mostafa2019dsr,
  title={Parameter Efficient Training of Deep Convolutional Neural Networks by Dynamic Sparse Reparameterization},
  author={Mostafa, Hesham and Wang, Xin},
  booktitle={Proceedings of the International Conference on Machine Learning (ICML)},
  year={2019},
  pages={4646--4655}
}

@inproceedings{cekic2022neuro,
  title={Neuro-Inspired Deep Neural Networks with Sparse, Strong Activations},
  author={Cekic, Mert and Bakiskan, Can and Madhow, Upamanyu},
  booktitle={Proceedings of the IEEE International Conference on Acoustics, Speech and Signal Processing (ICASSP)},
  year={2022}
}

@inproceedings{schwarz2021powerpropagation,
  title={Powerpropagation: A sparsity inducing weight reparameterization},
  author={Schwarz, Jonathan and others},
  booktitle={Advances in Neural Information Processing Systems (NeurIPS)},
  year={2021}
}

@inproceedings{morcos2018importance,
  title={On the importance of single directions for generalization},
  author={Morcos, Ari S and Barrett, David G.T. and Rabinowitz, Neil C. and Botvinick, Matthew},
  booktitle={International Conference on Learning Representations (ICLR)},
  year={2018}
}

@inproceedings{zenke2017synaptic,
  title={Continual learning through synaptic intelligence},
  author={Zenke, Friedemann and Poole, Ben and Ganguli, Surya},
  booktitle={Proceedings of the International Conference on Machine Learning (ICML)},
  year={2017}
}

@article{russakovsky2015imagenet,
  title={ImageNet Large Scale Visual Recognition Challenge},
  author={Russakovsky, Olga and Deng, Jia and Su, Hao and others},
  journal={International Journal of Computer Vision (IJCV)},
  year={2015},
  volume={115},
  number={3},
  pages={211--252}
}

@inproceedings{trockman2022patches,
  title={Patches Are All You Need?},
  author={Trockman, Asher and Kolter, Zico},
  booktitle={International Conference on Learning Representations (ICLR)},
  year={2022}
}

@inproceedings{farquhar2018towards,
  title={Towards Robust Evaluations of Continual Learning},
  author={Farquhar, Sebastian and Gal, Yarin},
  booktitle={NeurIPS Continual Learning Workshop},
  year={2018}
}

@inproceedings{lopez2017gradient,
  title={Gradient episodic memory for continual learning},
  author={Lopez-Paz, David and Ranzato, Marc'Aurelio},
  booktitle={Advances in Neural Information Processing Systems (NeurIPS)},
  year={2017}
}

@inproceedings{achille2018emergence,
  title={Emergence of Invariance and Disentanglement in Deep Representations},
  author={Achille, Alessandro and Soatto, Stefano},
  booktitle={International Conference on Learning Representations (ICLR)},
  year={2018}
}

@inproceedings{fernando2017pathnet,
  title={PathNet: Evolution Channels Gradient Descent in Super Neural Networks},
  author={Fernando, Chrisantha and et al.},
  booktitle={Proceedings of the International Conference on Machine Learning (ICML)},
  year={2017}
}

@inproceedings{tian2020contrastive,
  title={Contrastive Representation Distillation},
  author={Tian, Yonglong and Krishnan, Dilip and Isola, Phillip},
  booktitle={CVPR},
  year={2020}
}

@inproceedings{zhou22moe,
  title= {Mixture-of-Experts with Expert Choice Routing},
  author= {Zhou, Yujia and Lei, Tao and Liu, Han and Du, Nan and Huang, Yichong and Dai, Zihang and Eisner, Jason and Ahmed, Amr},
  booktitle = {Advances in Neural Information Processing Systems (NeurIPS)},
  year= {2022},
}
